\documentclass[lettersize,journal]{IEEEtran}
\usepackage{amsmath,amsfonts}
\usepackage{algorithm}
\usepackage{array}
\usepackage{textcomp}
\usepackage{stfloats}
\usepackage{url}
\usepackage{verbatim}
\usepackage{graphicx}
\usepackage{cite}
\usepackage{algorithm}
\usepackage[noend]{algpseudocode} % or omit [noend] if you want explicit \EndIf, etc.
\usepackage{glossaries}
\usepackage{subcaption}

\DeclareCaptionLabelFormat{closing}{#2)}

\usepackage{siunitx}
\newacronym{mra}{MRA}{Magnetic Resonance Angiography}
\newacronym{mri}{MRI}{Magnetic Resonance Imaging}
\newacronym{cta}{CTA}{Computed Tomography Angiography}
\newacronym{ct}{CT}{Computed Tomography}
\newacronym{rrt}{RRT}{Rapidly Exploring Random Tree}
\newacronym{lcca}{LCCA}{Left Common Carotid Artery}
\newacronym{rcca}{RCCA}{Right Common Carotid Artery}
\newacronym{bca}{BCA}{Brachiocephalic Artery}
\newacronym{lvo}{LVO}{Large Vessel Occlusion}
\newacronym{lsa}{LSA}{Left Subclavian Artery}
\newacronym{rsa}{RSA}{Right Subclavian Artery}
\newacronym{dofs}{DOFs}{Degrees Of Freedom}

\usepackage{xcolor}
\usepackage{etoolbox}
\usepackage{tikz}
\usepackage[normalem]{ulem}

\newtoggle{colorMode}

\newcommand{\myBlue}[1]{%
    \iftoggle{colorMode}{\textcolor{blue}{#1}}{#1}%
}
\newcommand{\myRed}[1]{%
  \iftoggle{colorMode}{\textcolor{red}{\sout{#1}}}{\unskip\ignorespaces}%
}

\newcommand{\myReview}[1]{%
    \iftoggle{colorMode}{%
        \colorbox{yellow}{\textbf{\textcolor{black}{#1}}}%
    }{}%
}

\togglefalse{colorMode}
\begin{document}

\title{Contact-aware Path Planning for Autonomous Neuroendovascular Navigation}

\author{Aabha Tamhankar$^1$, Ron Alterovitz$^2$, Ajit S. Puri$^3$, Giovanni Pittiglio$^1$
\thanks{$^1$ FuTURE Lab, Department of Robotics Engineering, Worcester Polytechnic Insitute (WPI), Worcester, MA 01605, USA. Email: {\tt\small \{astamhankar, gpittiglio\}@wpi.edu}\newline
$^2$ Department of Computer Science, University of North Carolina at Chapel Hill, Chapel Hill, NC 27599, USA. Email: {\tt\small ron@cs.unc.edu} \newline
$^3$ UMass Memorial Medical Center, Worcester, MA 01655, USA. Email: {\tt\small ajit.puri@umassmemorial.org}
}}



\maketitle

\begin{abstract}
We propose a deterministic and time-efficient contact-aware path planner for neurovascular navigation. The algorithm leverages information from pre- and intra-operative images of the vessels to navigate pre-bent passive tools, by intelligently predicting and exploiting interactions with the anatomy. A kinematic model is derived and employed by the sampling-based planner for tree expansion that utilizes simplified motion primitives. This approach enables fast computation of the feasible path, with negligible loss in accuracy, as demonstrated in diverse and representative anatomies of the vessels. In these anatomical demonstrators, the algorithm shows a 100\% convergence rate within 22.8s in the worst case, \myBlue{with sub-millimeter tracking errors ($< \SI{0.64}{\milli\meter}$)}, and is found effective on anatomical phantoms representative of $\sim$94\% of patients.
\end{abstract}


\IEEEpeerreviewmaketitle
\section{Introduction}
\label{sec:introduction}
Stroke is the second most common cause of death globally, causing 11.6\% of all fatalities in 2019 \cite{Rai2023UpdatedCare}. A stroke-related death occurs approximately every 3 minutes and 14 seconds \cite{StrokeFacts}, and half of the stroke survivors suffer from lasting disabilities \cite{Donkor2018}, including physical, cognitive, and speech impairments. Ischemic strokes, caused by a blood clot blocking blood flow, comprise 87\% of all stroke cases. Among them, \glspl{lvo} are ischemic incidents involving blockages in the brain's major arteries, impacting about 295,000 Americans annually \cite{StrokeFacts}. In these instances, mechanical thrombectomy is used to restore blood flow \cite{Lee2023}. 

This neuroendovascular procedure entails using pre-operative 3D images, such as \gls{cta} and/or \gls{mra}, and intra-operative fluoroscopy to guide concentric guidewires and catheters to the site of ischemia. Because of its complexity, this minimally invasive procedure is only performed by skilled neurointerventional teams. However, recent studies revealed that 20\% of patients in the USA lived over 1 hour away from medical centers capable of \gls{lvo} treatment \cite{Yu2021} and show discouraging rural-urban disparities \cite{Hammond2020}. Timely treatment is vital, as each hour of delay before thrombectomy increases the likelihood of disability \cite{Powers2019}. Such delays contribute to the observed 20\% higher stroke mortality rate in rural patients compared to urban counterparts \cite{Llanos-Leyton2022DisparitiesStroke}. 

%After stroke symptoms are identified in an emergency unit, \gls{ct} and/or \gls{mri} scans are used to determine the stroke type. If ischemia is suspected, patients are directed to specialized centers for a 3D scan of the vessels, typically an \gls{cta} and/or \gls{mra}, to examine the arterial layout from the aortic arch to the brain vessels \cite{vanderZijden2019}, and to strategize the insertion of flexible tools from either the femoral or radial artery to the obstruction site (Fig. \ref{fig:platform}). Highly trained interventional radiologists use concentric guidewires and catheters (Fig. \ref{fig:model}) to reach to the clot using pre-operative \gls{cta} or \gls{mra} images and real-time fluoroscopy and eventually remove it. 

%However, only select large medical facilities offer this training environment and workload to maintain clinicians' proficiency \cite{Hammond2020}. 

\begin{figure}
    \centering
    \includegraphics[width=\columnwidth]{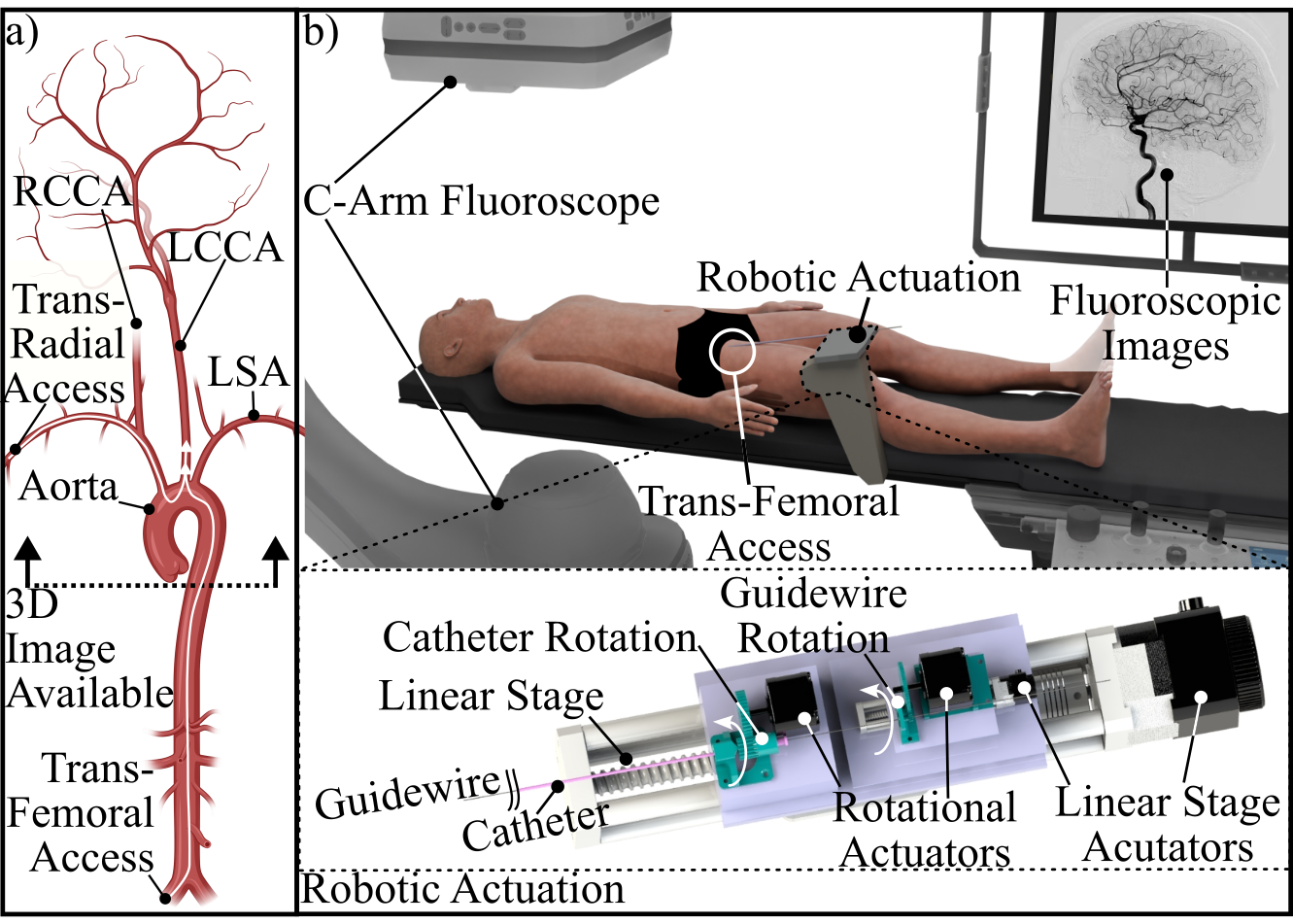}
    \caption{Description of autonomous neuroendovascular robotic solution. a) vessel anatomy; b) actuation platform. \myReview{R2-1}\myBlue{\textit{Reproduced from \cite{Tamhankar2025TowardsPlanning} \copyright 2025 IEEE.}}}
    \label{fig:platform}
\end{figure}

Our objective is to create an autonomous robotic system (Fig. \ref{fig:platform}) that utilizes pre-operative \gls{cta} or \gls{mra} scans for guiding pre-bent telescopic tools to the target. This would be used in smaller and rural hospitals to minimize time to treatment and ensure timely reperfusion. Medical staff would gain access to the femoral or radial artery and oversee tool navigation with guidance from remote expert surgeons in telecommunication. 

Under the premise that enhanced tip steerability can improve navigation efficiency, various actuation technologies such as cable- \cite{Lis2022, Abah2024Self-SteeringInterventions} and magnetic-based systems \cite{Kim2022, Dreyfus2024DexterousAccess, Brockdorff2024} have been explored for neuroendovascular navigation. These solutions, while possibly improving precision and targeting, entail high development, manufacturing expenses, and/or complications in their clinical translations; e.g. the need for large and expensive magnetic actuation units \cite{Kim2022, Dreyfus2024DexterousAccess, Brockdorff2024} or specific tendon-actuation designs \cite{Lis2022, Abah2024Self-SteeringInterventions}. Moreover, applications such as neurovascular interventions target long and complex anatomies which require control of too many independent \gls{dofs} to enable contact-free motions proposed for other endoluminal interventions \cite{Pittiglio2022, Pittiglio2023}.

Inspired by \cite{Chi2020CollaborativeLearning, Karstensen2023, Tamhankar2025TowardsPlanning}, we propose a robotic platform which can operate the pre-bend passive tools, currently used for stroke treatment. These tools, whose shape cannot be controlled in real-time, are inserted and rotated from their base outside the patient's body (Fig. \ref{fig:platform}) and navigate by only exploiting contact with the surrounding anatomy. This strategy, currently successful in clinical practice, does not require costly and complicated designs of the robotic system. However, to enable autonomy, integration of contact awareness is crucial for successful planning and navigation.

This paper proposes a deterministic and time-efficient contact-aware planner which, similarly to an expert neuroendovascular surgeon, is able to interpret pre- and intra-operative images and predict how interactions with the anatomy would enable navigating through the vessels. Drawing inspiration from previously proposed approaches \myReview{E1-5, R1-2} \myBlue{\cite{Zhang2024TowardCurvature, Tamhankar2025TowardsPlanning, Rao2024TowardsRobots, Du2025Physics-EmbeddedRobots, Wang2025Topology-InformedContacts}}, we consider deterministic modeling of robot-anatomy interactions. However, we study a simplified kinematics approach which, similarly to \cite{Greer2020RobustEnvironment}, considers a finite set of motion primitives observed in the interaction between tools and anatomy. This conversion of an iteratively-solved constrained optimization models approaches like \cite{Zhang2024TowardCurvature, Tamhankar2025TowardsPlanning, Rao2024TowardsRobots, Du2025Physics-EmbeddedRobots, Wang2025Topology-InformedContacts} into a simplified kinematic equivalent enables faster compute times with negligible loss in accuracy -- as shown in our experiments.

In this paper, similarly to \myReview{R1-3} \myBlue{\cite{Chi2020, Karstensen2023, Tamhankar2025TowardsPlanning}}, we focus on the first portion of the neuroendovascular navigation past the bifurcations between the aorta and the carotid arteries -- which eventually lead to the rest of the brain vessels. We consider the use of tools which are softer than the vessels and assume that the vasculature is static. However, given the fast computation of our planner, we predict its future use with intra-operative real-time imaging in the presence of anatomical motion.

\myReview{R1-1} We \myBlue{numerically} studied the efficiency and effectiveness of our approach \myBlue{through 100 independant trials} in the three most common aortic anatomies. \myRed{and} \myBlue{We} found that our algorithm achieves 100\% planning success within 52,000 iterations (approx. 22.8s) in the worst-case scenario -- 22.4$\times$ faster than \cite{Tamhankar2025TowardsPlanning}. \myRed{We present experimental results which show that, in the three diverse anatomies, we can autonomously navigate.}%
\myBlue{We experimentally validated the approach by integrating the planner into a robotic actuation platform to demonstrate autonomous navigation} to the \gls{rcca} and \gls{lcca}, of interest in most stroke treatments. We show successful navigation using localization of the tools based on registered 2D intra-operative fluoroscopy and 3D pre-operative \gls{cta} of anatomical features which represent approximately 94\% of patients.

\section{Time-Efficient Physics-Based Modeling Using Motion Primitives}
\label{sec:motionprimitives}
%We observed three fundamental motion primitives that define the behavior of the guidewire-catheter system during contact-aware navigation through vascular anatomy, as illustrated in Fig. \ref{fig:motionprimitives}.%: sliding on the surface of the vessel (Fig. \ref{fig:motionprimitives}a); using the wire to move past a bifurcation (Fig. \ref{fig:motionprimitives}b); use the catheter tip angle ($\theta$) to navigate a bifurcation (Fig. \ref{fig:motionprimitives}c). These primitives show how the tool interacts with vessel walls depending on local geometry and tool actuation.
We consider the problem of using a set of concentric guide-wires and catheters to access the base of the skull via the carotid arteries, past the aorta. Eventually, this allows access to the rest of the brain vessels. A standard wire used in this situation is a guidewire with a pre-bend tip of an angle \SI{45}-\SI{90}{\degree} and radius of curvature \SI{5}{\milli\meter}-\SI{15}{\milli\meter}. The catheter through which the wire is threaded has an angled tip at an angle $\theta$, typically between \SI{30}{\degree} and \SI{90}{\degree}. 
To allow navigation past the bifurcations between the aortic arch and carotid arteries (Fig. \ref{fig:platform}), these tools of fixed and passive tip bending are pushed to the boundary constraints created by the vessels' walls. %Our goal is to create a time-efficient planner which, in contrast with the traditional obstacle-avoidant approaches, is able to intelligently utilize constraints for navigation -- similarly to a neuroendovascular surgeon. 

We use pre-operative 3D imaging of the vessels registered to real-time fluoroscopy to predict contact-enabled navigation past the bifurcations. These imaging modalities, typically available for these procedures, allow localizing our tools within the environment, as in our experimental analysis in Section \ref{sec:experiments}. In this work, we assume that the tools can be considered much softer than the anatomy and that the latter is static. 

\begin{figure}
    \centering
    
    \includegraphics[scale=0.71]{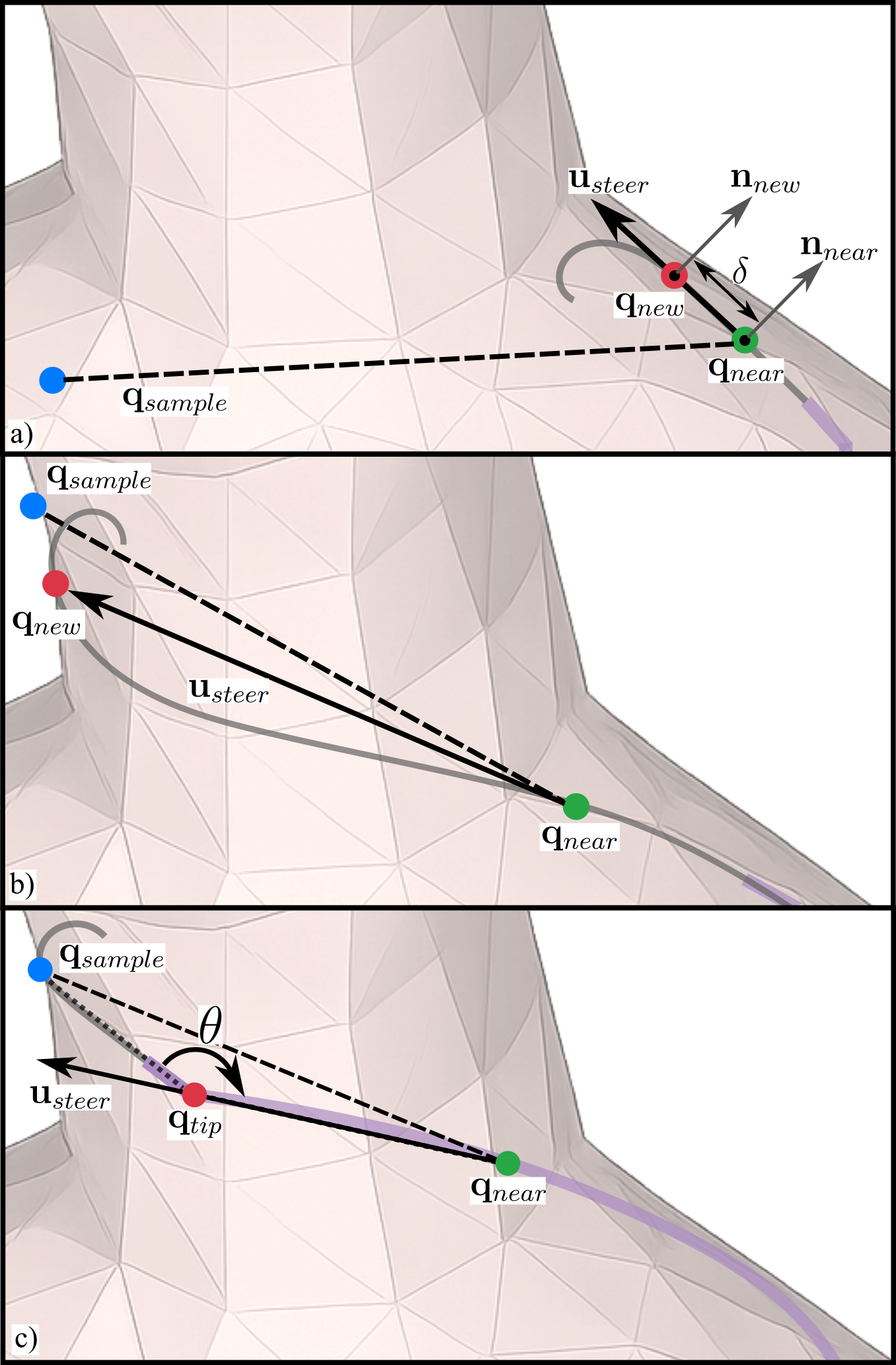}
    
    \caption{\myReview{E1-6, R1-6} Schematic of the three motion primitives in our kinematic planner: 
(a) \textit{wall-guided glide}, with the guidewire (dark gray) sliding along the vessel wall; 
(b) \textit{free-space rebound}, where the guidewire traverses a bifurcation and re-contacts the opposite wall; 
(c) \textit{angled-catheter launch}, where the guidewire exits the catheter (purple) at a fixed bend angle.} 
    \label{fig:motionprimitives}
\end{figure}

%To obtain a plan based on tool-anatomy interaction, one is required to model such interaction. The approach considered in previous works \cite{Zhang2024TowardCurvature, Tamhankar2025TowardsPlanning, Rao2024TowardsRobots} is to formulate the statics of the tools and use the environment as constraints. However, this requires solving a generally time-inefficient constrained optimization. 
To enable timely computation of the plan, we formulate the problem in the kinematics domain, avoiding solving the constrained statics as proposed in previews works \cite{Zhang2024TowardCurvature, Tamhankar2025TowardsPlanning, Rao2024TowardsRobots}. This kinematic formulation is based on the observation that the motion of the wire and catheter can be simplified by three motion primitives: a) \textit{wall-guided glide}, where the wire slides along the convex portions of the anatomy, naturally adapting to the environment (Fig. \ref{fig:motionprimitives}a); b) \textit{free-space rebound}, where the wire travels across free space between two walls, which is observed in concave portions of the anatomy (Fig. \ref{fig:motionprimitives}b); c) \textit{angled-catheter launch}, where the wire extends from the catheter at the characteristic catheter angle $\theta$ (Fig. \ref{fig:motionprimitives}c). Unlike the free-flight motion in b), where the wire drifts until it makes incidental contact with the opposite wall, the angled-catheter launch provides a controlled exit direction. This directed motion allows the wire to access vessel walls or branch points that would otherwise be unreachable through passive free-space travel alone. %While in practice the wire may lose its pre-shaped configuration if pushed forcefully against a fixed point, thereby altering the angle θ due to its flexibility, this scenario is not considered in our formulation. Instead, the planner assumes that the wire retains its intended shape throughout navigation. We assume that wires that undergo permanent deformation can be substituted with an alternative to ensure consistent and predictable behavior.

The proposed contact-aware planner, detailed in Section \ref{sec:planning}, utilizes these motion primitives to create time-efficient connections between each new sample and other nodes in the tree. Eventually, the plan is converted into base insertion and rotation of the tools, via time-efficient geometric inversion of the kinematics, as described in Section \ref{sec:ik}.

\section{Contact-aware Path Planning}
\label{sec:planning}
From pre-operative imaging modalities such as \gls{cta} or \gls{mra}, we extract a triangulated surface mesh  representing the patient-specific vascular anatomy. Our goal is to compute a feasible navigation path from a user-defined start point $\mathbf{q}_{\text{init}} \in \mathbb{R}^3$ to one or more goal locations on the vessel surface. We define the goal region ($\mathcal{X}_{\text{goal}}$) as the set of mesh triangles corresponding to the distal tips of target branches, which were manually annotated. The path is constructed such that each node lies on the vessel surface and corresponds to a configuration that the guidewire or catheter can reach via wall contact, consistent with the motion primitives described in Section~\ref{sec:motionprimitives}.  %$\mathcal{M}$

\myReview{R2-2} \myBlue{The proposed surface-constrained planning framework is summarized in Algorithm~\ref{alg:planner}.} Each node $\mathbf{q}  \in \mathbb{R}^3$ in the tree is a point, the on a face of the triangular mesh, associated with a normal vector $\mathbf{n} \in \mathbb{R}^3$, and the corresponding tangent plane. Our planner proceeds by incrementally building a tree $\mathcal{T}$ in the space of surface-contact points.

At each iteration, a random point $\mathbf{q}_{\text{sample}}$ is sampled on the mesh using uniform surface sampling. The closest node $\mathbf{q}_{\text{near}}$ in the tree is selected based on Euclidean distance, i.e.

\begin{equation}
\mathbf{q}_{\text{near}} = \arg\min_{\mathbf{q} \in \mathcal{T}} \| \mathbf{q} - \mathbf{q}_{\text{sample}} \|.
\end{equation}

We then compute the steer vector $\mathbf{u}_{\text{steer}}$, i.e. the direction of motion from from $\mathbf q_{\text{near}}$ to $\mathbf q_{\text{sample}}$, projected onto the tangent plane of $\mathbf q_{\text{near}}$’s triangle 
\begin{equation}
\mathbf{u}_{\text{steer}} = (\mathbf q_{\text{sample}} - \mathbf q_{\text{near}}) - \left( \left( \mathbf q_{\text{sample}} - \mathbf q_{\text{near}} \right)\cdot\mathbf{n}_{\text{near}} \right)\ \mathbf{n}_{\text{near}},
\end{equation}
where $\mathbf{n}_\text{near}$ is the unit normal of the triangle face at $\mathbf q_{\text{near}}$. This projection ensures that steering respects surface contact, as the wire is assumed to slide along the vessel wall unless redirected.

% These steer vector is used to then to towards and form a new node. Here we have three options to select a new node give the steer vector, and this related to the motion primitives given in Section \ref{sec:motionprimitives}.  
To ensure feasibility, the planner checks that the angle between $\mathbf{u}_{\text{steer}}$ and the vector connecting $\mathbf q_{\text{sample}}$ with its parent node $\mathbf q_{\text{previous}}$ is within the threshold $\tau_\text{bend}$, i.e.
%Next, the planner checks whether this steering vector respects a bend threshold $\tau_{\text{bend}}$ using \texttt{computeAlignment}, ensuring that only physically plausible steering motions are allowed. Let $\mathbf{v}_{\text{prev}}$ be the unit vector along the previous path segment. The alignment metric is computed as
% \begin{equation}
% \alpha = \mathbf{v}_{\text{prev}} \cdot {\mathbf{u}}_{\text{steer}},
% \end{equation}
% and the motion is accepted only if
\begin{equation}
\label{eq:threshold}
\mathbf{u}_{\text{steer}} \cdot \left(\mathbf q_{\text{near}} - \mathbf q_{\text{previous}}\right) \leq \tau_{\text{bend}},
\end{equation}
If (\ref{eq:threshold}) is respected, $\mathbf q_{\text{sample}}$ is added to the tree $\mathcal{T}$, and the connection between $\mathbf q_{\text{sample}}$ and $\mathbf q_{\text{near}}$ has to be determined. Depending on the concavity of the anatomy at $\mathbf q_{\text{near}}$, the connection falls under one of the primitives in Fig. \ref{fig:motionprimitives}. 

% To determine whether the anaotmical region around the nearest node is geometrically concave or convex, we use \texttt{checkConcavity}, which returns a scalar curvature heuristic $\kappa$. If $\kappa > 0$, the region is locally concave, corresponding to the "wall-guided glide" motion primitive as illustrated in Fig \ref{fig:motionprimitives}a. In this case, the tool is assumed to maintain surface contact. A candidate node is generated by stepping a fixed distance $\delta$ along $\mathbf{u}_{\text{steer}}$ and projected back onto the mesh via \texttt{projectToClosestSurface} to get $q_{\text{new}}$.

The function \texttt{checkConcavity} returns a scalar curvature heuristic $\kappa$ computed from the surface normals of the current triangle and the triangle adjacent to $\mathbf q_{\text{near}}$ in the direction of $\mathbf{u}_{\text{steer}}$ 
%Let $\mathbf{n}_{\text{near}}$ be the unit normal vector of the triangle containing $q_{\text{near}}$, and let $\mathbf{n}_{\text{adj}}$ be the unit normal vector of the adjacent triangle in the heading direction given by $\mathbf{u}_{\text{steer}}$. The concavity measure is defined as
\begin{equation}
\kappa = \mathbf{u}_{\text{steer}} \cdot (\mathbf{n}_{\text{near}} \times \mathbf{n}_{\text{adj}}),
\end{equation}
with $\mathbf{n}_{\text{adj}}$ is the normal of the adjacent triangle to $\mathbf{q}_{\text{near}}$ in the direction $\mathbf{u}_{\text{steer}}$. 

The sign of $\kappa$ indicates the concavity.
When $\kappa > 0$, the region is locally concave, corresponding to the \textit{wall-guided glide} motion primitive (Fig.~\ref{fig:motionprimitives}a). In this case, the tool maintains surface contact, and a candidate node is generated by moving a fixed distance $\delta$ along $\mathbf{u}_{\text{steer}}$. This target is then projected back onto the mesh surface using \texttt{projectToClosestSurface} to obtain $\mathbf q_{\text{new}}$.

If the region is convex  ($\kappa < 0$), the tool loses wall contact and enters free space until it re-establishes contact. In this case, we implement the two motion primitives illustrated in Fig.~\ref{fig:motionprimitives}b and Fig.~\ref{fig:motionprimitives}c.

The first option is a \textit{angled-catheter launch} as in Fig.~\ref{fig:motionprimitives}c. The catheter tip at $\mathbf q_{\text{near}}$ acts as a base, and the guidewire emerges at a fixed bend angle $\theta$ relative to the catheter axis. We model this by defining a virtual tip point
\begin{equation}
\mathbf q_{\text{tip}} = \mathbf q_{\text{near}} + \lambda \, \mathbf{u}_{\text{steer}}, \quad \lambda > 0,
\end{equation}
where $\mathbf{u}_{\text{steer}}$ is the projected steering vector. The point $\mathbf q_{\text{tip}}$ is chosen such that the geometric constraint
\begin{equation}
\angle \big( \mathbf q_{\text{sample}} - \mathbf q_{\text{tip}}, \;\mathbf q_{\text{near}} - \mathbf q_{\text{tip}} \big) = \theta
\end{equation}
is satisfied. This is implemented in \texttt{useAngledCatheter}, which simulates the guidewire leaving the catheter at a controlled angle. If $\mathbf q_{\text{tip}}$ lies within the anatomical mesh, a ray is cast from $\mathbf q_{\text{tip}}$ toward $\mathbf q_{\text{sample}}$; the first valid intersection with the vessel wall defines the new node $\mathbf q_{\text{new}}$.

If the catheter-based launch is not feasible (e.g., $\mathbf q_{\text{tip}}$ lies outside the anatomy or no valid intersection is found), the planner falls back to a \textit{free-space rebound}. In this case, a ray is cast directly from $\mathbf q_{\text{near}}$ along $\mathbf{u}_{\text{steer}}$, and the first valid surface intersection becomes the new node $\mathbf q_{\text{new}}$. This simulates the wire advancing freely across the vessel cross-section until it makes contact again, consistent with the behavior illustrated in Fig.~\ref{fig:motionprimitives}b.

\begin{algorithm}
\caption{RRT-Based Catheter Path Planning on Vascular Surface Meshes}
\label{alg:planner}
\begin{algorithmic}[1]
\small
\Require Vascular triangle mesh $\mathcal{M}$, initial configuration $q_{\text{init}}$, step size $\delta$
\Require Flexibility threshold $\tau_{\text{bend}}$
\Require Maximum iterations $N_{\max}$

\State Initialize tree $\mathcal{T} \leftarrow \{\mathbf{q}_{\text{init}}\}$

\For{$i = 1$ to $N_{\max}$}

    \State $\mathbf{q}_{\text{sample}} \leftarrow$ \texttt{samplePointOnAnatomy}($\mathcal{M}$)
    \State $\mathbf{q}_{\text{near}} \leftarrow$ \texttt{findNearestNode}($\mathcal{T}$, $\mathbf{q}_{\text{sample}}$)
    \State $\mathbf{u}_{\text{steer}} \leftarrow$ \texttt{projectOnSurface}($\mathbf{q}_{\text{sample}} - \mathbf{q}_{\text{near}}$, $\mathcal{M}$)

    \If{$\texttt{computeAlignment}(\mathbf{u}_{\text{steer}}, \mathbf{q}_{\text{near}}) \ge \tau_{\text{bend}}$}
        \State $\kappa \leftarrow$ \texttt{checkConcavity}($\mathbf{u}_{\text{steer}}$, $\mathbf{q}_{\text{near}}$, $\mathcal{M}$)

        \If{$\kappa > 0$}
            \State $\mathbf{q}_{\text{target}} \leftarrow \mathbf{q}_{\text{near}} + \mathbf{u}_{\text{steer}} \cdot \delta$
            \State $\mathbf{q}_{\text{new}} \leftarrow$ \texttt{projectToClosestSurface}($\mathbf{q}_{\text{target}}$, $\mathcal{M}$)
        \Else
            \State $\mathbf{q}_{\text{new}} \leftarrow$ \texttt{computeRayIntersection}($\mathbf{q}_{\text{near}}$, $\mathbf{u}_{\text{steer}}$, $\mathcal{M}$)
            \State $\mathbf{q}_{\text{tip}} \leftarrow$ \texttt{angledCatheter}($\mathbf{q}_{\text{near}}$, $\mathbf{u}_{\text{steer}}$)

            \If{\texttt{insideAnatomy}($\mathbf{q}_{\text{tip}}$, $\mathcal{M}$)}
                \State $\mathbf{q}_{\text{new}} \leftarrow$ \texttt{computeRayIntersection}($\mathbf{q}_{\text{tip}}$, $\mathbf{q}_{\text{sample}} - \mathbf{q}_{\text{tip}}$, $\mathcal{M}$)
            \Else
                \State $\mathbf{q}_{\text{new}} \leftarrow$ \texttt{computeRayIntersection}($\mathbf{q}_{\text{near}}$, $\mathbf{u}_{\text{steer}}$, $\mathcal{M}$)
            \EndIf
        \EndIf

        \State \texttt{addTreeNode}($\mathcal{T}$, $\mathbf{q}_{\text{near}}$, $\mathbf{q}_{\text{new}}$)
    \EndIf

\EndFor

\If{$\mathbf{q}_{\text{new}} \in \mathcal{X}_{\text{goal}}$}
    \State path $\leftarrow$ \texttt{extractPath}($\mathcal{T}, \mathbf{q}_{\text{new}}$)
    \State \Return path, $\mathcal{T}$
\EndIf

\end{algorithmic}
\end{algorithm}

A successful plan is the first feasible set of surface-constrained connected nodes  from the start node to the target region using a combination of the three motion primitives. From the resulting tree $\mathcal{T}$, we extract the path from $\mathbf{q}_{\text{init}}$ to the reached goal point. This path can be executed by our robotic platform which actuates insertion and rotation of both guidewire and catheter, as depicted in Fig. \ref{fig:platform}.

%%%%%%%%%%%%%%%%%%%

\section{Inverse Kinematics}
\label{sec:ik}
The planner described in Section~\ref{sec:planning} produces a surface-constrained 3D trajectory on the anatomical mesh. Each point along this trajectory corresponds to a location on the vessel wall where the guidewire is expected to make contact before advancing to the next. To execute this path on our robotic platform, we calculate control inputs for the catheter–guidewire system, which supports four degrees of motion: insertion and rotation of the catheter, and insertion and rotation of the guidewire.

%The catheter is first used to follow the initial portion of the trajectory. Once the planner employs the catheter-based launch primitive (Fig.~\ref{fig:motionprimitives}c), control is switched to the guidewire, which then completes the navigation to the target. The planner is designed to use the catheter-based primitive only once, and the control is handed over to the guidewire for the remainder of the navigation.

\subsection{Guidewire-only Motion Primitives}
When the connection between $\mathbf q_\text{near}$ and $\mathbf q_\text{new}$ falls under the guidewire motion primitives in Figs.  \ref{fig:motionprimitives}a-b, we compute the amount of insertion of the guidewire, i.e. 
\begin{equation}
d_g = \| \mathbf q_\text{new} - \mathbf q_\text{near} \|.
\end{equation}
The axial rotation of the guidewire is controlled so that the bent portion points away from the vessel's surface, sliding along the anatomy. We define the tangent vector at the tip of the wire $\mathbf t_g$ and the normal at $\mathbf q_\text{new}$ as $\mathbf n_\text{new}$; the rotation of the guidewire $\phi_g$ is found as 
\begin{equation}
\label{eq:wire_rotation}
    \phi_g = \arctan\left(\frac{\mathbf t_g \cdot \mathbf n_\text{new}}{\|\mathbf t_g \times \mathbf n_\text{new}\|}\right)
\end{equation}

\subsection{Guidewire and Catheter Motion Primitive}
Actuation of the catheter is primarily used in the third motion primitive in Fig. \ref{fig:motionprimitives}c, where we first consider the insertion from  $\mathbf q_\text{near}$ to $\mathbf q_\text{tip}$, i.e. 

\begin{equation}
d_c = \| \mathbf q_\text{tip} - \mathbf q_\text{near} \|
\end{equation}
and the insertion of the guidewire is found as 
\begin{equation}
d_g = \| \mathbf q_\text{sample} - \mathbf q_\text{tip} \|.
\end{equation}

Guidewire rotation $\phi_g$ is found as in (\ref{eq:wire_rotation}) and the catheter is controlled so that its tip points to $\mathbf{q}_\text{sample}$. We define $\mathbf{t}_c$ as the tangent at the catheter tip, $\mathbf{s} = \mathbf q_\text{sample} - \mathbf q_\text{tip}$ and 

\begin{equation}
    \phi_c = \arctan\left(\frac{\mathbf t_c \cdot \mathbf s}{\|\mathbf t_c \times \mathbf s\|}\right).
\end{equation}

Using this approach, we convert the plan in Section \ref{sec:planning} into a series of commands in the robot's configuration space. Eventually, these are executed as discussed in Section \ref{sec:experiments}.

% To steer the tool along the trajectory, the bend direction of the catheter/guidewire tip should be aligned with the path direction. We define the \textit{bend vector} $\mathbf{b}$ as the vector from the start of the bent segment to the tool tip. For two successive path segments, $\mathbf{v}_1 = q_{i} - q_{i-1}$ and $\mathbf{v}_2 = q_{i+1} - q_i$, we project both vectors and $\mathbf{b}$ onto the tangent plane of the vessel surface at $q_i$, defined by the local surface normal $\mathbf{n}_i$. The projected vectors are
% \begin{equation}
% \mathbf{a} = \mathbf{v}_1 - \frac{\mathbf{v}_1 \cdot \mathbf{n}_i}{\|\mathbf{n}_i\|^2} \mathbf{n}_i, 
% \quad
% \mathbf{b} = \mathbf{v}_2 - \frac{\mathbf{v}_2 \cdot \mathbf{n}_i}{\|\mathbf{n}_i\|^2} \mathbf{n}_i.
% \end{equation}
% The signed rotation angle between $\mathbf{a}$ and $\mathbf{b}$ is
% \begin{equation}
% \phi_i = \text{atan2}\!\big( (\mathbf{a} \times \mathbf{b}) \cdot \mathbf{n}_i , \; \mathbf{a} \cdot \mathbf{b} \big),
% \end{equation}
% which is used as the rotation command to align the bend vector with the trajectory direction.

% Each trajectory step is thus translated into a pair of commands: an insertion $d_i$ followed by a rotation $\phi_i$. Thus, we convert the surface constrained trajectory into a sequence of alternating \texttt{translate}-\texttt{rotate} commands, which are written to a control file for execution. 

% This conversion assumes ideal transmission, i.e., base actuation translates directly to the tool tip and friction is assumed negligible.

\section{Results}
\label{sec:results}
In this section, we present the validation of the Contact-Aware \gls{rrt} planner explained in Section \ref{sec:planning} on anatomies of the aortic arch. The primary objective is to guide a combination of guidewire and catheter from the base of the descending aorta to the \gls{lcca} and \gls{rcca} via the \gls{bca}, while avoiding the \gls{lsa} and any other branches arising from the aortic arch. These may include anatomical variants such as \gls{rsa}, an aberrant right subclavian artery (ARSA),or a directly originating vertebral artery, if present.

We use the publicly available Aortic Vessel Tree (AVT) \gls{cta} Dataset and Segmentations \cite{Radl2022AorticSegmentations}, which consists of \gls{cta} scans from predominantly healthy subjects. These scans cover the aortic arch and its branches, as well as the abdominal aorta and iliac arteries. Binary masks for the aortas and their branches are provided as part of the dataset, enabling the required anatomical segmentations.

\begin{figure}
    \centering
    \includegraphics[width=\columnwidth]{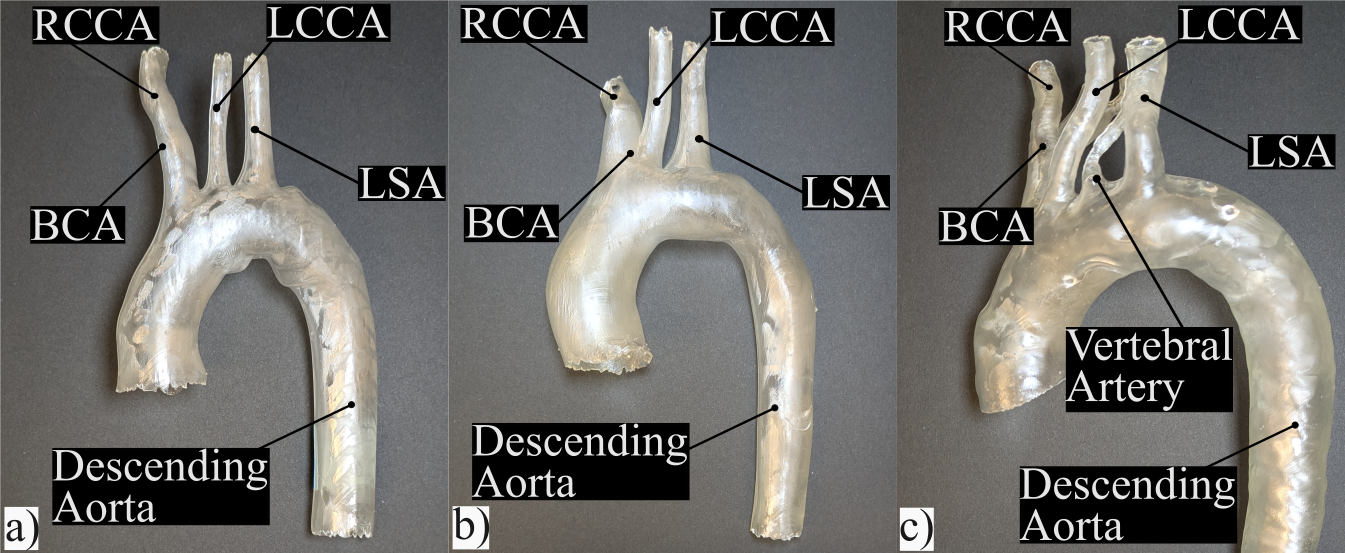}
    \caption{3D printed models of three aortic arch variants: a) Type I (classic), b) Type II (bovine), and c) Type III (direct vertebral origin).} 
    \label{fig:aa_variants}
\end{figure}

From this dataset, we selected three distinct aortic arch anatomies: classic, bovine, and direct vertebral origin (Type I, II, and III, respectively), as illustrated in Fig. \ref{fig:aa_variants}. These variants represent the most common aortic arch anatomies, constituting approximately 94\% of the patients. The classic aortic arch (Fig \ref{fig:aa_variants}a) is the most common configuration, where the \gls{bca} arises first, followed by the \gls{lcca}, and then the \gls{lsa}. This configuration is found in approximately 65-94\% of individuals. The bovine arch (Fig. \ref{fig:aa_variants}b) is a variant where the BCA and LCCA share a common origin. This anatomical variation occurs in about 11-27\% of the population. The direct vertebral origin variant (Fig. \ref{fig:aa_variants}c) is relatively rare, occurring in 2.5-8\% of individuals, where the vertebral artery originates directly from the aortic arch rather than from the subclavian artery \cite{Natsis2009AnatomicalReview,Shapiro2018AorticCatheterization}. We used these three anatomies for our numerical and experimental validations.

All algorithms were implemented in MATLAB and executed on a standard laptop-class CPU (Intel Core i5, 3.2\,GHz, 8\,GB RAM). Results are presented in two parts: (i) numerical validation on aortic arch anatomies segmented from \gls{cta} data, and (ii) experimental validation \myBlue{of 3D navigation} \myReview{R1-5} in 3D-printed anatomical models under fluoroscopic guidance.

\subsection{Numerical Validation}
\label{sec:num_ana}

The three anatomies described in Section~\ref{sec:results} (see Fig. \ref{fig:aa_variants}) were segmented to isolate aortic arch and its supra-aortic branches, and converted into mesh representations as inputs for the contact-aware planning described in Section \ref{sec:planning}.

For each anatomy, 100 start points were randomly generated at the descending aorta for the guidewire and catheter system. Given that both the catheter and guidewire are passive, a wire initially travels along free space until it contacts the vessel wall. Hence, the start node for the planner was defined as the first point of contact along the wire’s axis on the anatomy surface. The targets were defined as the endpoints of the LCCA and RCCA branches for each anatomy.

Each iteration of our proposed planner required an average of $0.439\,\text{ms}$. Under identical hardware and implementation conditions, we benchmarked the approach in \cite{Tamhankar2025TowardsPlanning}, which required $10.7\,\text{ms}$ per iteration, approximately $22.4\times$ slower. In addition, the runtime of our planner scales linearly with the number of iterations, since both nearest-neighbor search and ray–triangle intersection are implemented with linear-time algorithms.

\begin{figure}
    \centering
    \includegraphics[width = \columnwidth]{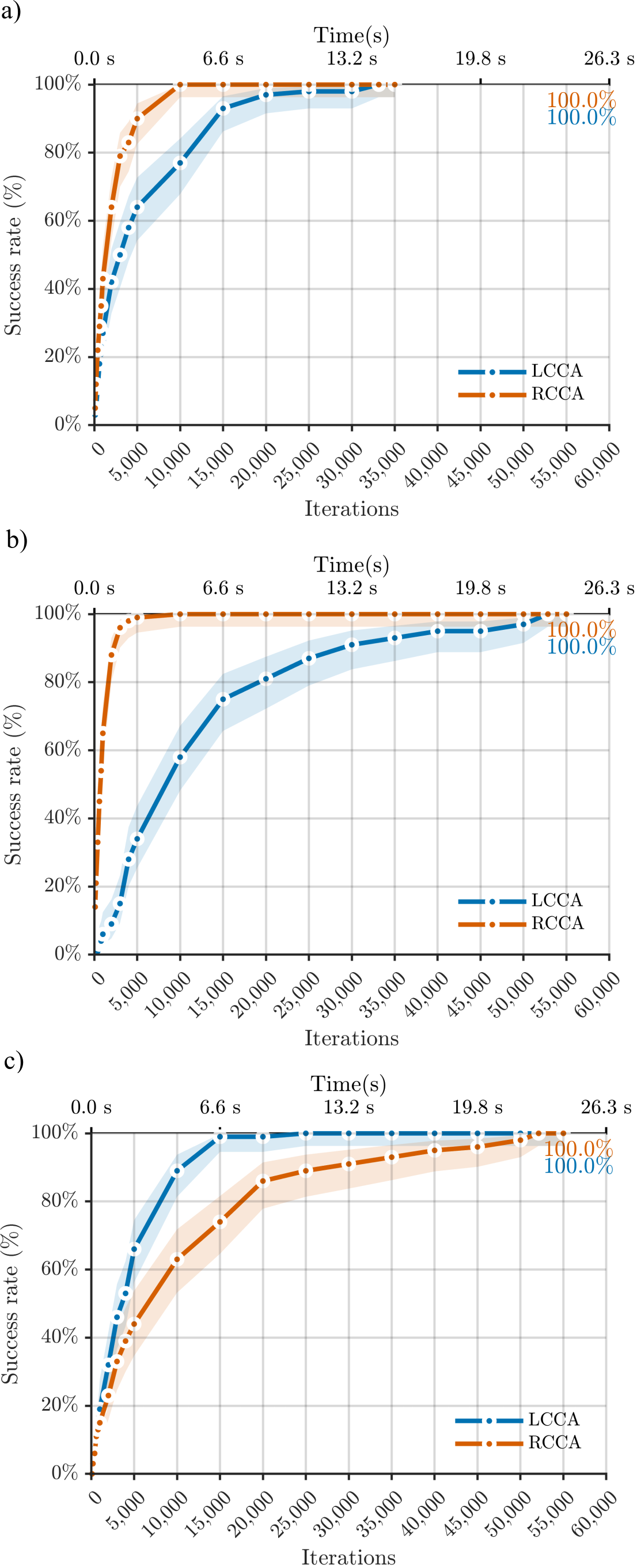}
    \caption{Success rate versus number of iterations for three aortic arch variants: (a) classic, (b) bovine, and (c) direct vertebral origin.} 
    \label{fig:num_ana}
\end{figure}

The planner was executed for 60,000 iterations per anatomy. Figure \ref{fig:num_ana} shows the relationship between the number of iterations and the empirical success rate of reaching the targets. The X-axis denotes the number of iterations  while the Y-axis shows the probability of successfully reaching the \gls{lcca} and \gls{rcca}. Each curve reflects the probability of success across 100 randomized trials, with shaded regions indicating 95\% Wilson confidence intervals.

\myReview{R2-3}In the classic arch, both targets are reached with high reliability within relatively few iterations. The success rate curves rise steeply, with \myBlue{RCCA} achieving 100\% success by approximately \myBlue{10,000} iterations or \myBlue{4.4s}, and \myBlue{LCCA} reaching 100\% success by 25,000 iterations or 11.0s (Fig. \ref{fig:num_ana}a). The classic anatomy provides the most direct pathways with minimal geometric constraints.

In bovine arch, the success curves exhibit more gradual growth, reflecting the increased geometric complexity. RCCA achieves 100\% success more rapidly (10,000 iterations, 4.4s), while LCCA requires \myBlue{more than 50,000 iterations} to reach 100\% success (Fig. \ref{fig:num_ana}). The shared origin of the \gls{lcca} and brachiocephalic artery requires more samples for the planner to discover valid trajectories to \gls{lcca}.

The direct vertebral origin anatomy presents the most challenging navigation scenario. LCCA achieves 100\% success \myBlue{(around 25,000 iterations, 11s)}, while RCCA exhibits a more gradual convergence, requiring approximately \myBlue{40,000 iterations (17.6s)} to reach saturation near 95\% (Fig. \ref{fig:num_ana}).

The results demonstrate the probabilistic completeness property of RRT-based planning; success probability increases monotonically with iteration budget and asymptotically approaches saturation. However, convergence rates are strongly anatomy-dependent, with more complex branching patterns requiring substantially more iterations to achieve practical success rates.
All anatomies eventually achieve success rates of 100\% within a maximum of 22.8s, validating the contact-aware planning approach across diverse geometric configurations.

% Across all variants, the observed curves reflect the probabilistic completeness property of RRT; the success probability increases monotonically with the iteration budget and asymptotically approaches saturation. However, the iterations required to achieve practical success rates is anatomy-dependent. 

% Each mesh used in our experiments contained approximately 700–1000 faces. On a standard laptop-class CPU (3–4 GHz), one iteration of Algorithm \ref{alg:catheter_rrt} required an average of 0.439ms time, with the exact value depending on the current tree size. Since both the nearest-neighbor search and the ray–triangle intersection are currently implemented with linear scans, runtime scales linearly with the number of faces and nodes.

\myBlue{
\subsection{Failure Analysis}
\label{sec:failure}
\myReview{E1-3}
\myBlue{

\begin{figure}
    \includegraphics[scale=0.72]{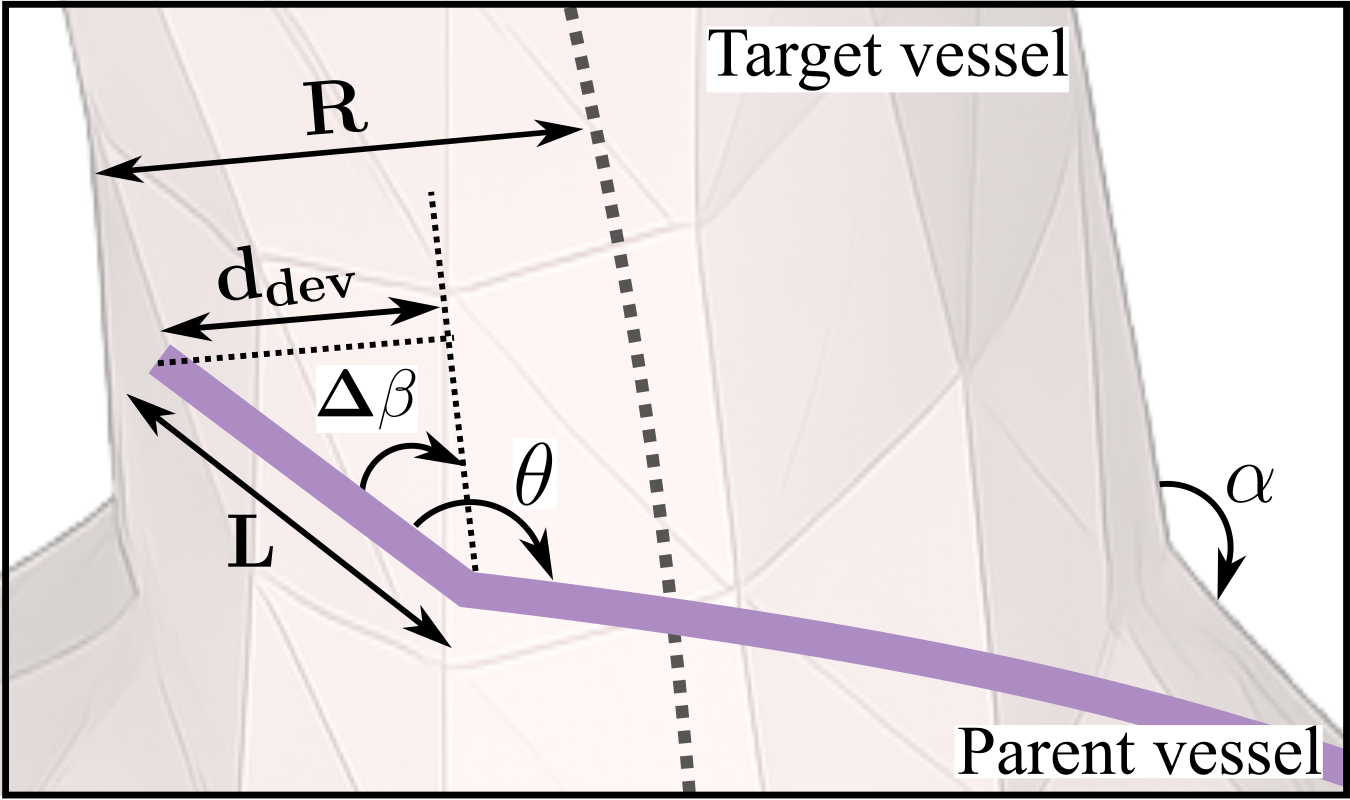}
    \caption{Geometric parametrization of failure conditions for the \textit{angled-catheter launch}.}
    \label{fig:failure}
\end{figure}

Among the motion primitives described in Section \ref{sec:planning}
, the \textit{angled-catheter launch} is critical for accessing branch vessels that exhibit tortuous or non-tangential takeoffs. Hence, it is neccessary to derive the geometric boundaries of this primitive and formulate the failure conditions as a function of vessel topology and the kinematic constraints of the catheter-guidewire system.

Consider a bifurcation where a parent vessel branches into a target vessel as illustrated in Fig. \ref{fig:failure}. We model the target branch as a cylindrical tube with radius $R$. We define the \textit{branch takeoff angle} $\alpha \in [0, \pi]$ as the angle at which the branch takes off from the parent vessel. The distal catheter tip has a fixed bend angle $\theta$ relative to its longitudinal axis. Let $L$ denote the effective bent tip length, defined as the Euclidean distance from the bend vertex to the distal tip of the catheter.

In an ideal continuous configuration space, one would select a tool such that $\theta = \alpha$, resulting in perfect coaxial alignment of the tool with the target vessel branch. However, in discrete path planning with a finite set of surgical tools, angular mismatch is inevitable.

We define the angular misalignment $\Delta \beta$ as the absolute difference between the anatomical takeoff angle and the catheter bend angle:
\begin{equation}
    \Delta \beta = | \alpha - \theta |.
\end{equation}

As the tool traverses the tip length $L$, this angular error propagates into a lateral displacement orthogonal to the target branch centerline. The lateral deviation $d_{\text{dev}}$ is given by:
\begin{equation}
    d_{\text{dev}} = L \sin(\Delta \beta) = L \sin(| \alpha - \theta |).
\end{equation}

A geometric failure occurs if this lateral deviation exceeds the vessel radius, causing the wire to impact the vessel wall rather than entering the branch. Thus, the failure condition is:
\begin{equation}
    \label{eq:catheter_failure}
    R < L \cdot \sin(| \alpha - \theta |).
\end{equation}

This inequality reveals a fundamental coupling between geometry and tool selection: as the standoff distance $L$ increases, the allowable angular error decreases. Consequently, successful navigation of distant or small-caliber branches ($R \to 0$) requires a catheter inventory with high angular resolution.

To validate the practical applicability of the proposed failure bounds, we compare theoretical constraints against aortic arch morphology, where vessel takeoff angles range from standard Type I ($\alpha \in [65^\circ, 90^\circ]$) to complex Type III ($\alpha \in [25^\circ, 45^\circ]$) configurations \cite{Demertzis2010AorticHumans}. Standard commercial catheter libraries offer a discrete spectrum of pre-formed shapes—from straight tips ($\theta \approx 15^\circ$) to recurved geometries ($\theta \approx 120^\circ$)—that effectively span these angular domains. By selecting a tool from this set that minimizes the angular misalignment $\Delta \phi$, the planner can be utilized to guarantee reachability in realistic anatomies.
}

}

\subsection{Experimental Validation}
\label{sec:experiments}

\begin{figure}
    \includegraphics[width=\columnwidth]{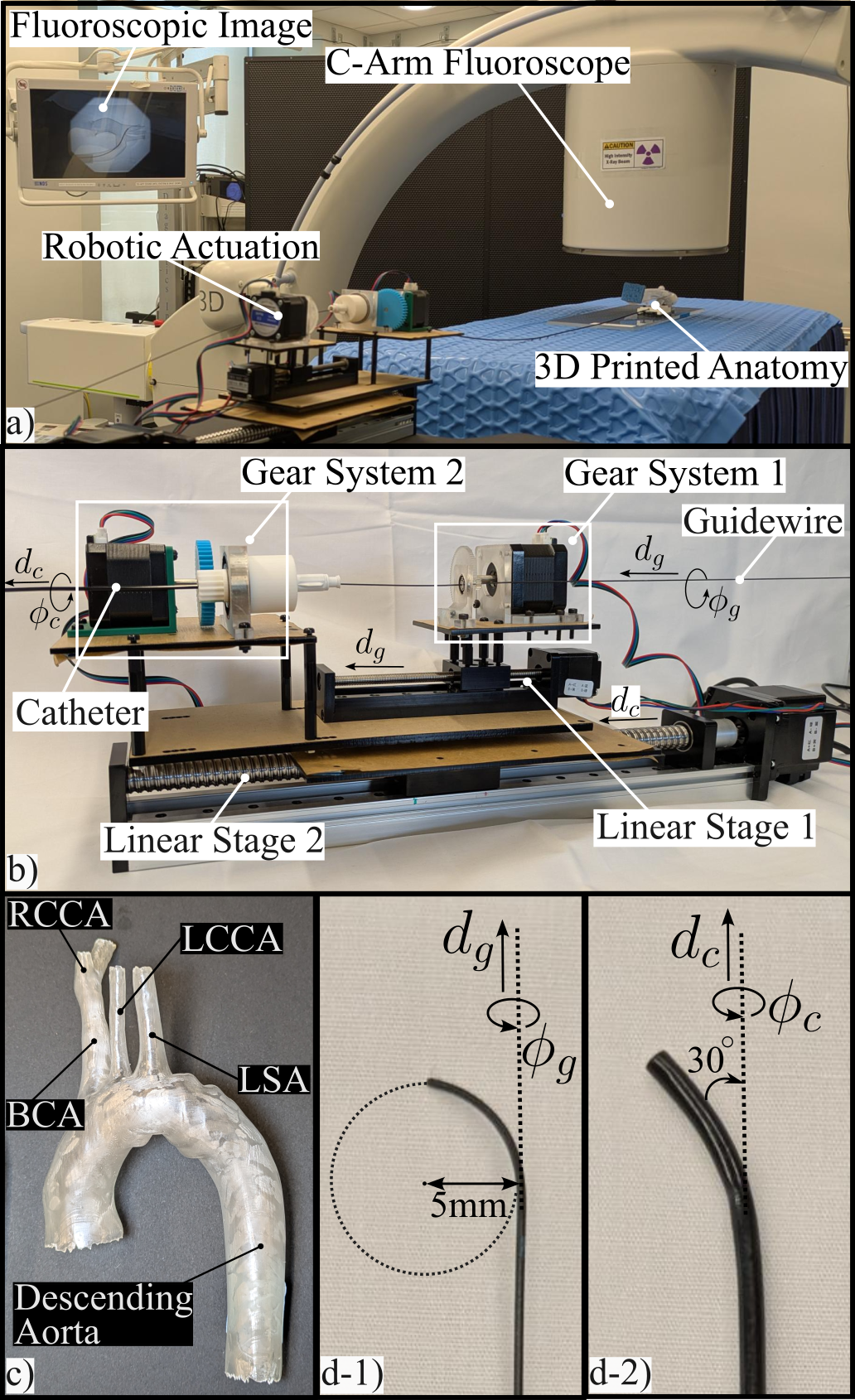}
    \caption{Experimental Setup: a) setup, b) robotic actuation system, c) 3D printed anatomy, d-1) guidewire  and d-2) catheter.} 
    \label{fig:actuationsys}
\end{figure}

For the experimental demonstration of our method, the three anatomies described in Section~\ref{sec:results} (see Fig. \ref{fig:aa_variants}), segmented from medical images, were fabricated using stereolithography (SLA) 3D printing. A fluoroscopic imaging setup (Fig.~\ref{fig:actuationsys}a) was employed to capture anteroposterior (A/P) projections of the 3D-printed models, consistent with standard clinical practice during neuroendovascular procedures. The C-arm configuration enabled real-time visualization of the anatomical model and the guidewire-catheter system during navigation.

For these experiments, Radifocus Glidewire Guidewire with a \SI{5}{\milli\meter} radius of curvature (Fig. \ref{fig:actuationsys} d-1), along with the Asahi Fubuki Neurovascular Sheath featuring a \SI{30}{\degree} angled tip (Fig. \ref{fig:actuationsys} d-2), were used. \myReview{E1-1} \myBlue{Both the tools are manufactured to be coated with a hydrophilic layer designed to reduce friction, allowing them to slide smoothly along the anatomy. This aligns with the mechanics of our planner, which relies on wall contact to guide the tool toward the target.} These tools, aligned coaxially in a flushed configuration, were positioned at the base of the descending aorta. Due to gravity, the wire naturally rested along the inferior vessel wall. %, simplifying registration in a single fluoroscopic plane. 
Anatomical landmarks, specifically the endpoints of the ascending and descending aorta, were used to co-register the fluoroscopic frame with the 3D model used by the planning algorithm. The initial position of the guidewire and catheter was found using such registration and used as the start configuration for our planner.  %In clinical settings, these landmarks could include key features easily identifiable by clinicians, such as the base of any major branch or the tip of the ascending aorta, which would allow for efficient registration during procedures.

For each anatomy, navigation trials were conducted \myBlue{in 3D} \myReview{R1-5} targeting both LCCA and RCCA endpoints. The experimental protocol consisted of path planning using the contact-aware RRT algorithm (Section \ref{sec:planning}), motion command generation through the inverse kinematics solver (Section \ref{sec:ik}), robotic execution under continuous fluoroscopic monitoring, and trajectory tracking through sequential frame capture.

%Given that both the catheter and guidewire are passive, if the wire is pushed along its axis, it will initially travel through free space until it makes contact with the vessel wall. Therefore, the start node of the planner was defined as the point where the guidewire, when pushed along its axis from its original position, would first make contact with the wall.

A custom 4-DOF robotic actuation system (Fig.~\ref{fig:actuationsys}b) was developed to execute motion commands generated by the inverse kinematics framework described in Section~\ref{sec:ik}. The platform enables independent translation and rotation control for both the catheter and guidewire. The guidewire translation is controlled by Linear Stage 1, its rotation is managed by Gear System 1, while the catheter translation is achieved using Linear Stage 2, its rotation is controlled by Gear System 2.

\begin{figure}[t]
    \centering
    
    % --- Subfigure (a) ---
    \begin{subfigure}{0.48\textwidth}
        \centering
        \includegraphics[width=\linewidth]{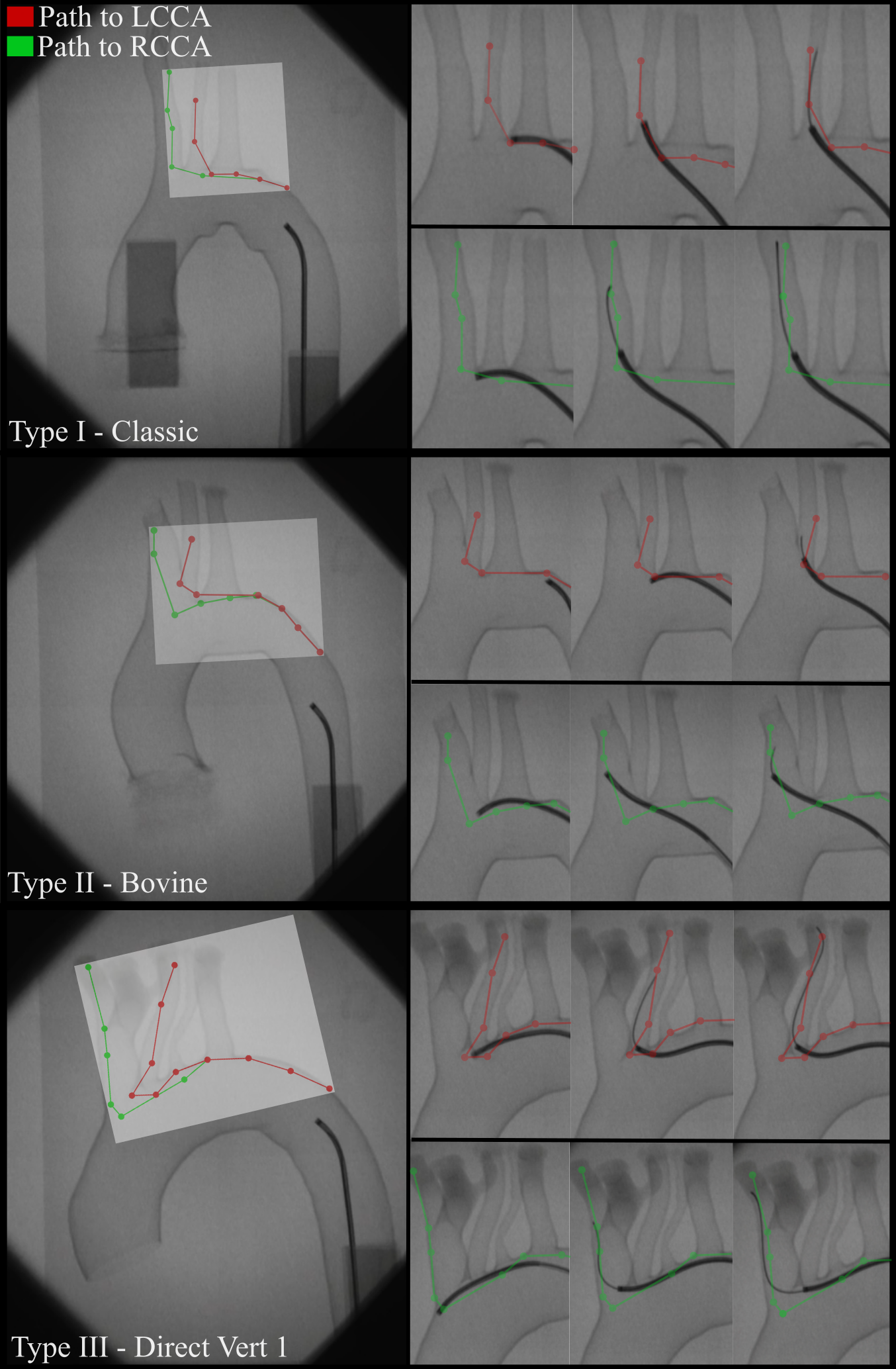}
        \caption{Navigation under fluoroscopy for three aortic arch types.}
        \label{fig:results_a}
    \end{subfigure}
    \hfill
    % --- Subfigure (b) ---
    \begin{subfigure}{0.48\textwidth}
        \centering
        \includegraphics[width=\linewidth]{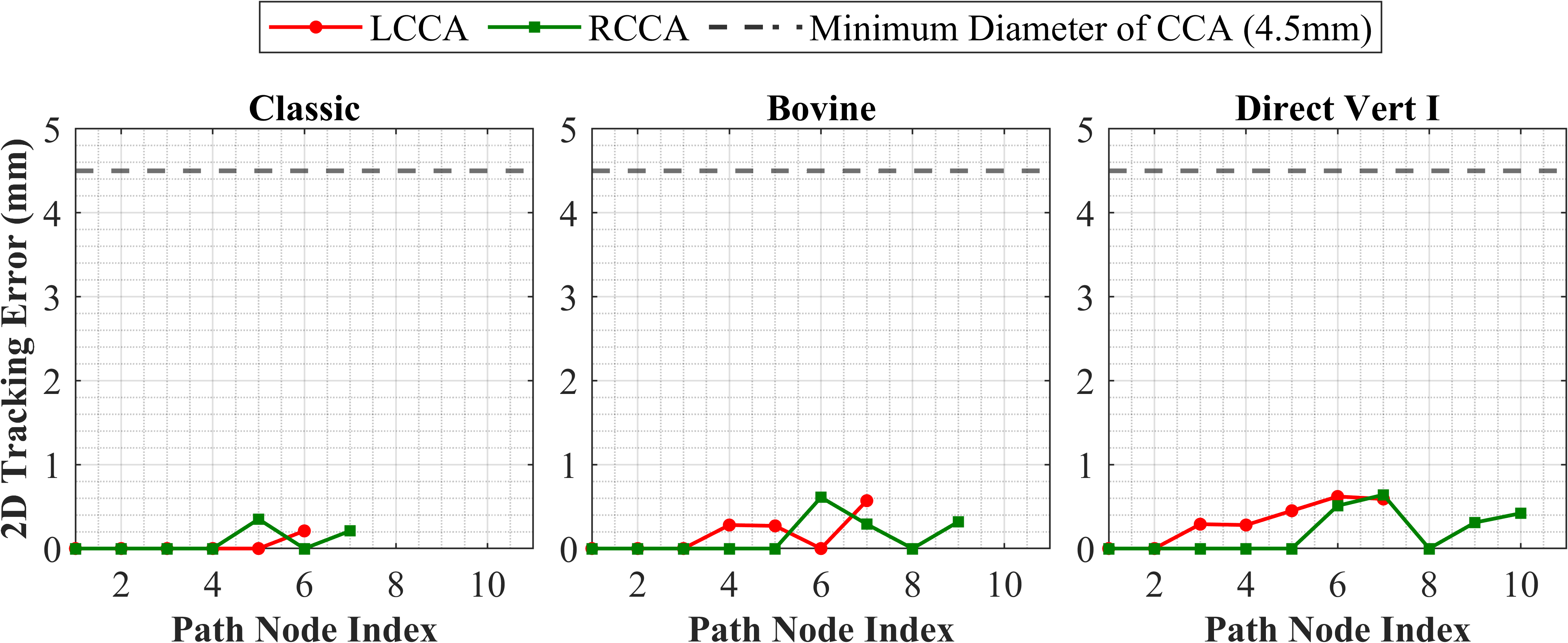}
        \caption{Navigation tracking error.}
        \label{fig:results_b}
    \end{subfigure}

    \caption{Experimental results.}
    \label{fig:results}
\end{figure}

The experimental results are summarized in Fig.~\ref{fig:results}. \myBlue{To visualize the trajectory, the generated 3D path is projected onto the 2D fluoroscopic plane.} In all anatomies, the guidewire followed the surface-constrained paths generated by the planner when advanced under fluoroscopic guidance. \myRed{Sequential fluoroscopy frames were captured during navigation, enabling comparison between the planned trajectory and the actual guidewire position within the 3D-printed aortic models. The frames where extracted from continuous fluoroscopic scan shown in the Supplementary Video.} \myReview{E1-2, R2-4} \myBlue{To quantify execution accuracy, sequential frames were extracted from the continuous fluoroscopic scan (see Supplementary Video). We projected the planner's discrete 3D surface contact nodes onto the calibrated 2D imaging plane to serve as ground truth targets. Figure~\ref{fig:results_b} details the tracking error, defined as the Euclidean distance in the image plane between each projected target node and the actual physical guidewire position. The analysis covers navigation tasks to both the LCCA and RCCA across the Classic, Bovine, and Direct Vertebral aortic arch models. The system demonstrated a maximum recorded tracking error of \SI{0.64}{\milli\meter}. Given that the minimum luminal diameter of the adult Common Carotid Artery is approximately \SI{4.5}{\milli\meter} \cite{Krejza2006CarotidSize}, this sub-millimeter deviation confirms that the robotic system can successfully execute the planner's surface-constrained motion primitives while maintaining safe margins within the vessel.}

The results confirmed that the incremental translation and rotation commands reliably advanced the guidewire along the vessel wall toward the target branches (LCCA and RCCA). Across trials, the executed trajectories qualitatively matched the pre-planned paths, demonstrating the ability of the proposed framework to translate geometric motion primitives into experimentally realizable guidance.

% \myBlue{The minimum statistical CCA luminal diameter is approximately 4.5 mm in adults, as established in the study by \cite{Krejza2006CarotidSize}.}

\section{Conclusions}
\label{sec:conclusions}
The present paper discusses a computationally-efficient contact-aware planner for neuroendovascular procedures. The planner is novel in nature for its ability to exploit anatomical contact to navigate through the anatomy. Our method leverages pre-operative 3D scans of the vessels registered to intra-operative real-time 2D images to predict feasible paths enabled by interacting with the vessels' walls.

The proposed planner leverages kinematic primitives requiring minimal computation time, compared to previous approaches that require iterative solutions of complex optimization problems. This is fundamental in vascular procedures, specifically when time equals brain loss such as for stroke.

We present a comprehensive numerical analysis which shows that the method achieves 100\% accuracy within 22.8 s in the worst-case scenario, in three geometrically distinct aortic arches representative of 94\% of patients. Experiments performed with our robotic platform in the same three anatomies also show successful navigation, \myReview{E1-1} \myRed{proving that the proposed kinematic planner can be successfully deployed in realistic clinical settings.} \myBlue{with sub-millimeter tracking errors ($< \SI{0.64}{\milli\meter}$). We attribute these deviations to unmodeled dynamics, including friction, mechanical play, and non-linear transmission losses between actuation at the base and the distal tip. While the open-loop execution proved sufficient for these phantom studies given the safety margins, future work will focus on implementing closed-loop control strategies to actively compensate for these physical discrepancies in dynamic, real-world clinical environments.}

\bibliographystyle{IEEEtran}
\bibliography{references, stroke_bib}	

\end{document}